\documentclass[letterpaper, 10 pt, conference]{ieeeconf}  

\IEEEoverridecommandlockouts                              

\usepackage[ruled,linesnumbered]{algorithm2e}
\SetKwRepeat{Do}{do}{while}

\usepackage[hyphens]{url}
\usepackage{hyperref}
\newenvironment{myitemize}{\begin{list}{$\bullet$}
{\setlength{\topsep}{1mm}
\setlength{\itemsep}{0.25mm}
\setlength{\parsep}{0.25mm}
\setlength{\itemindent}{0mm}
\setlength{\partopsep}{0mm}
\setlength{\labelwidth}{15mm}
\setlength{\leftmargin}{4mm}}}{\end{list}}

\usepackage{array}
\newcommand{\PreserveBackslash}[1]{\let\temp=\\#1\let\\=\temp}
\newcolumntype{C}[1]{>{\PreserveBackslash\centering}p{#1}}
\newcolumntype{R}[1]{>{\PreserveBackslash\raggedleft}p{#1}}
\newcolumntype{L}[1]{>{\PreserveBackslash\raggedright}p{#1}}

\usepackage{graphics} 
\usepackage{epsfig} 
\usepackage{mathptmx} 
\usepackage{times} 
\usepackage{amsmath} 
\usepackage{amssymb}  
\usepackage{multirow}
\usepackage{bm}
\usepackage{graphicx}
\usepackage{xcolor}
\usepackage{diagbox}
\DeclareMathOperator*{\argmax}{arg\,max}

\newtheorem{theorem}{Theorem}[section]

\title{\LARGE \bf Safety-Assured Speculative Planning with Adaptive Prediction}

\author{Xiangguo Liu$^{1}$, Ruochen Jiao$^{1}$, Yixuan Wang$^{1}$, Yimin Han$^{1}$, Bowen Zheng$^{2}$, Qi Zhu$^{1}$
\thanks{$^{1}$Xiangguo Liu, Ruochen Jiao, Yixuan Wang, Yimin Han and Qi Zhu are with the Department of Electrical and Computer Engineering, Northwestern University, Evanston, IL 60208, USA. Emails:
        {\tt\small xg.liu@u.northwestern.edu, ruochen.jiao@u.northwestern.edu, YixuanWang2024@u.northwestern.edu, yiminhan2020@u.northwestern.edu, qzhu@northwestern.edu.}}%
\thanks{$^{2}$Bowen Zheng is with Pony.ai, Inc., Fremont, CA 94538, USA.
        Email: {\tt\small bowen.zheng@pony.ai.}}%
}

\begin{document}

\maketitle
\thispagestyle{empty}
\pagestyle{empty}

\begin{abstract}
Recently significant progress has been made in vehicle prediction and planning algorithms for autonomous driving. However, it remains quite challenging for an autonomous vehicle to plan its trajectory in complex scenarios when it is difficult to accurately predict its surrounding vehicles' behaviors and trajectories. In this work, to maximize performance while ensuring safety, we propose a novel speculative planning framework based on a prediction-planning interface that quantifies both the behavior-level and trajectory-level uncertainties of surrounding vehicles. Our framework leverages recent prediction algorithms that can provide one or more possible behaviors and trajectories of the surrounding vehicles with probability estimation. It adapts those predictions based on the latest system states and traffic environment, and conducts planning to maximize the expected reward of the ego vehicle by considering the probabilistic predictions of all scenarios and ensure system safety by ruling out actions that may be unsafe in worst case. We demonstrate the effectiveness of our approach in improving system performance and ensuring system safety over other baseline methods, via extensive simulations in SUMO on a challenging multi-lane highway lane-changing case study.
\end{abstract}

\section{Introduction}\label{sec: intro}




Autonomous driving has shown great promise to revolutionize the transportation system by improving its safety~\cite{riedmaier2020survey,lee2020regulations,sinha2020comprehensive,jiao2021end,chang2023safety,liu2023safe}
and performance~\cite{guo2019urban,montanaro2019towards,guo2019joint,stogios2019simulating,chen2023mixed}.
Extensive research has been conducted to improve the perception, prediction, and planning modules of individual vehicles for autonomous driving. However, some driving tasks and scenarios still remain quite challenging~\cite{cao2020reinforcement,liu2022physics,zhan2019interaction,jiao2022semi}, 
especially when there are complex interactions~\cite{liu2023safety} between the ego vehicle and its surrounding vehicles and it is difficult for the ego vehicle to accurately predict the behavior of its neighbors~\cite{jiao2023learning}. 

While connectivity technology~\cite{jin2022dsrc} has the potential to greatly mitigate the challenges in predicting the intention and future trajectory of surrounding vehicles~\cite{liu2022markov,liu2021securing,liu2023connectivity,liu2023waving}, it is expected that there will be a long transition period before the full deployment of connected vehicles and traffic infrastructures~\cite{luo2021credibility}. Moreover, in mixed traffic, human drivers have different driving patterns, which can even change over time~\cite{yu2018human}, while autonomous vehicles designed by different companies can have varied driving strategies for similar scenarios~\cite{huang2022compatibility}. All these make it critically important in autonomous driving to accurately predict the surrounding vehicles' behavior \emph{and} effectively leverage the prediction results in planning. 



\begin{figure}[tbp]
\centering\includegraphics[scale=0.45]{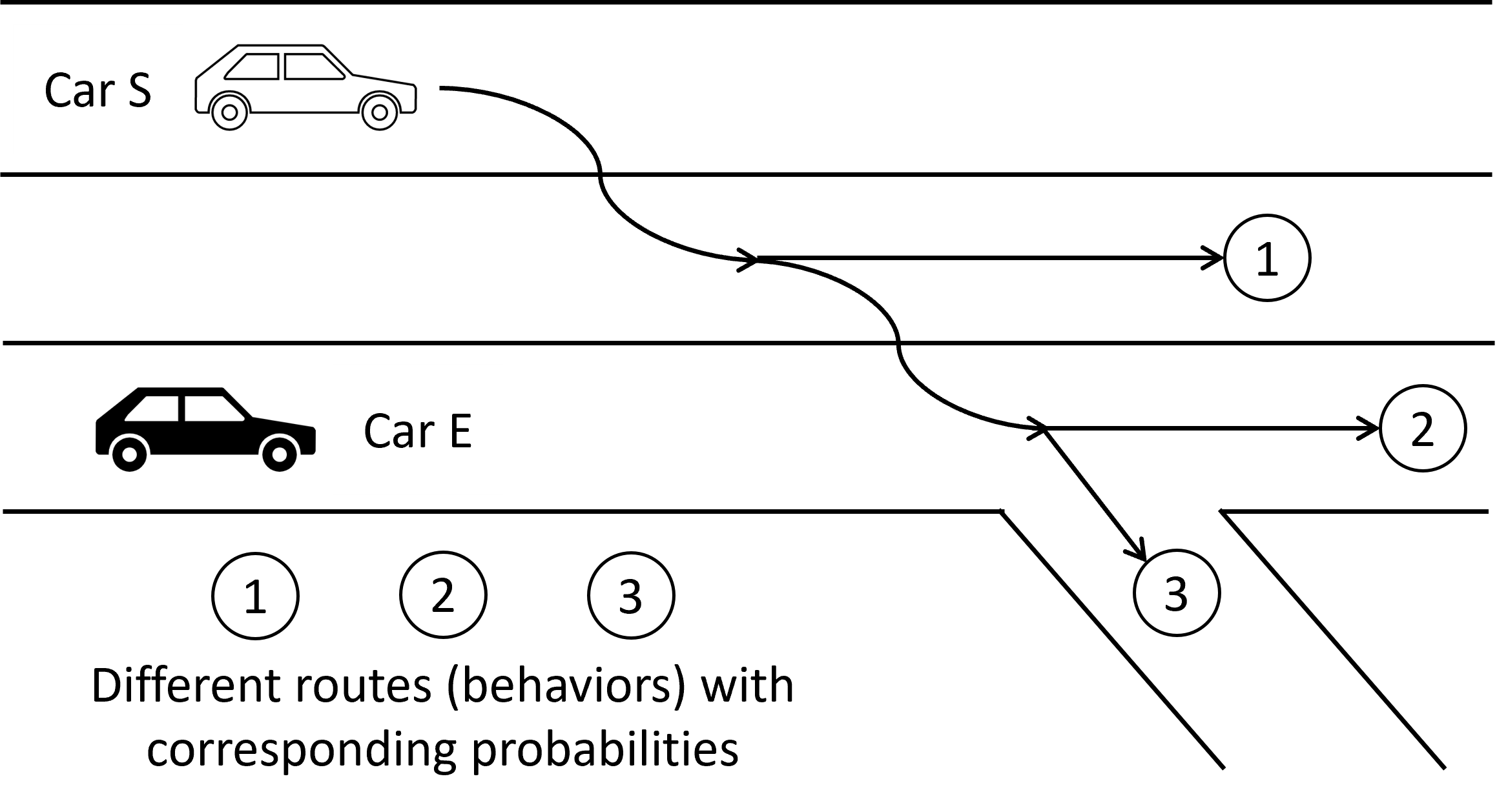}
\caption{Representative case study: In a multi-lane highway, an ego vehicle $E$ goes straight and a surrounding vehicle $S$ indicates a right turn to change lanes. However, the intention of $S$ is ambiguous. It can change lane once and follow the route $1$, or change lanes twice and follow the route $2$, or exit the highway after lane changing and follow the route $3$.}
\label{fig:scenario}
\end{figure}

Our work focuses on addressing complex driving scenarios where the surrounding vehicles' intentions and planned trajectories have multiple possibilities in prediction. 
For example, we consider the lane changing in a multi-lane highway as a representative application in this work (while our proposed approach can be extended to other driving tasks and scenarios). As shown in Fig.~\ref{fig:scenario}, the system includes an ego vehicle $E$ going straight and a surrounding vehicle $S$ indicating a right turn to change lanes. However, the intention of the surrounding vehicle is ambiguous. It can change lane once and follow the route $1$, or change lanes twice and follow the route $2$, or exit the highway after lane changing and follow the route $3$. Each possible route may be associated with a probability, i.e., $p_1$, $p_2$ and $p_3$. 
Moreover, besides the behavior-level uncertainty on taking which route, the exact parameters for defining each trajectory are also uncertain.


In the literature, some recent prediction algorithms can provide one or more most possible trajectories of a surrounding vehicle, albeit does not emphasize their behavior-level difference~\cite{cui2019multimodal,liang2020learning,zhao2021tnt,gu2021densetnt}. Some planning algorithms are designed to prevent traffic accidents for the most possible predicted trajectory, but ignore other possibilities and cannot guarantee safety in those cases~\cite{luo2016dynamic,li2021prediction}. There are other planning strategies that consider all possible predicted trajectories of a surrounding vehicle. However, it would be over-conservative if the most cautious action is always selected in considering all possibilities~\cite{zhan2016non}. Directly taking a `weighted' action across all possibilities can also be risky~\cite{liu2022physics,cao2020reinforcement} -- for example, when the traffic signal in an intersection just turns yellow, it may be safe for a vehicle to maintain its velocity and pass the intersection before the traffic signal turns red \emph{or} decelerate and stop before the intersection, but unsafe for it to take a weighted action like hesitating and entering the intersection at a low speed. 





To overcome these challenges, we propose a safety-assured speculative planning framework with adaptive prediction. The framework leverages prediction algorithms that can provide one or more possible behaviors and future trajectories of surrounding vehicles~\cite{sun2022domain, zhao2021tnt, liu2021multimodal}. During planning, the framework considers all those possibilities and first rules out the actions that may be unsafe in the worst case. Within the remaining actions, it selects the one that maximizes the expected reward (representing system performance) of all possible intentions and trajectories in prediction results, with larger weights assigned to the more likely ones. We consider such planning \emph{speculative} because the prediction results are likely to change over time. Thus, our framework also checks the updated prediction results over time and \emph{adapts} them based on system states and traffic environment, to filter out those impossible behaviors and trajectories of the surrounding vehicles for more effective planning. Moreover, we incorporate the prediction of the aggressiveness level of the surrounding vehicles into our prediction-planning interface, which may further reduce prediction uncertainty and improve system performance.


In summary, the contributions of our work include:
\begin{myitemize}
\item We propose a speculative planning method to address the challenges in ambiguous scenarios where multiple behaviors and trajectories of surrounding vehicles exist. Our method considers all possible predicted behaviors and trajectories, ensures system safety by ruling out actions that may be unsafe in the worst case, and improves system performance by sampling all possibilities and choosing the action that maximizes the expected reward. 
\item Our planner leverages a prediction-planning interface that incorporates uncertainty on both behavior level and trajectory level, including the probability distribution of relevant parameters. It reacts to the prediction changes in real time and adapts the prediction results based on the system states and traffic environment, to filter out impossible behaviors and trajectories of surrounding vehicles as time goes by for further improving system performance. 
\item We demonstrate the advantages of our approach over baseline methods in improving system performance and ensuring system safety, through extensive simulations under various scenarios. Note that our approach is guaranteed to be safe if the prediction results are conservative and there exists a safe planning decision at the initial state.
\end{myitemize}

The rest of the paper is organized as follows. In Section~\ref{sec: related work}, we review related works on prediction and planning in challenging scenarios. In Section~\ref{sec: methodology}, we present our method of speculative planning with adaptive prediction. Section~\ref{sec: experiment} shows the experimental results and Section~\ref{sec: conclusion} concludes the paper.


\section{Related Work}\label{sec: related work}
With the wide adoption of machine learning-based techniques~\cite{zhu2021safety}, the performance of trajectory prediction has significantly improved over the last several years. Most works predict the trajectories of traffic participants and evaluate their accuracy~\cite{jiao2022tae,liang2020learning,liu2021multimodal,ye2021tpcn}. There are also some works that consider both high-level behaviors and low-level trajectories in the prediction algorithms. For instance, \cite{xue2022integrated} proposes an integrated lane change prediction model to predict the lane change decisions and lane change trajectories. \cite{hu2022causal} develops a domain generalization method for prediction in unseen scenarios, and mainly works on behavior prediction. \cite{rhinehart2019precog} performs both standard forecasting and the novel task of conditional forecasting, which reasons about how all agents will likely respond to the goal of a controlled agent. It points out that goal/intent-conditioned trajectory forecasting can improve joint-agent and per-agent predictions, compared to unconditional forecast. \cite{sun2022domain} proposes a framework that first explicitly predicts the distribution of an agent’s endpoint over a discretized goal set and then completes the trajectories conditioned on the selected goal points. The goal set is designed with domain knowledge.

Planning has been a popular research topic for many years in areas including robotics, automotive engineering, transportation and autonomous driving. Various planner designs are reviewed in~\cite{gonzalez2015review,paden2016survey,claussmann2019review,zhu2021survey}. These planners can be classified according to the inputs and assumptions. Most of these planners such as ~\cite{luo2016dynamic,chen2017constrained} take the latest system states and environmental information as inputs, e.g., position and velocity of surrounding vehicles, road geometry, and so on. Based on the position and velocity of surrounding vehicles, it is usually assumed that vehicles keep the same velocity in the next few seconds~\cite{liu2023interactive}, thus the trajectory is acquired. Thus it is appropriate and convenient to take more accurately predicted trajectories as inputs to these planners~\cite{luo2016dynamic,liu2021trajectory}. There are also many planners with embedding models for surrounding vehicles, which describe the reward function with a set of parameters~\cite{yu2018human,liu2020impact} or neural networks~\cite{zhu2018human,xie2019data}. These methods can model the interaction between the ego vehicle and surrounding vehicles, and the future trajectories of these agents are derived at the same time. 

Significant progress has been made recently to reduce the uncertainty and safety risks during interactions among vehicles. Some works propose that it is safer and more effective to conduct motion planning with pre-determined or predicted behavior of surrounding vehicles~\cite{liu2022physics,wang2019cooperative,cao2020reinforcement}. Confidence-based methods~\cite{fisac2018probabilistically,tian2022safety} are also promising in human-robot interaction, where robots are designed to use confidence-aware game theoretic models of human behavior when assessing the safety~\cite{tian2022safety}. Confidence gets updated after comparing real human behavior and its predictions, and safety is ensured by switching to a safe planner when necessary. \cite{li2021prediction} integrates a probabilistic prediction model into the design of reachability-based safety controllers to achieve more efficient two-car collision avoidance. However, it cannot ensure safety because it only reacts to the most possible predicted trajectories of surrounding vehicles. \cite{huang2023risk} introduces a Bayesian Long Short-term Memory (BLSTM) model to predict the probability distribution of surrounding vehicles’ positions, which are used to estimate dynamic conflict risks. Model Predictive Control is then incorporated to navigate vehicles through safe paths with the least predicted conflict risk. However, it only considers trajectory-level uncertainty, and it could be over-conservative because collision risk is the only term for optimization.

Some prediction methods provide more information to quantify the uncertainty, which can improve the performance of planners in uncertain and dynamic environments. For instance, \cite{chai2019multipath} uses Gaussian Mixture Model to describe the predicted intentions, detailed waypoints, and corresponding covariance for each waypoint. \cite{cui2019multimodal,liang2020learning} work on multimodal trajectory predictions, and produce every trajectory's probability by a prediction network. \cite{phan2020covernet} predicts multimodal potential trajectories with corresponding probabilities based on a set of anchors. \cite{zhao2021tnt,gu2021densetnt} first predict possible goals for surrounding vehicles and then generate multiple detailed future trajectories. Our planner is compatible with a general interface, which can include both behavior-level and trajectory-level uncertainties.

\section{Our Proposed Framework}
\label{sec: methodology}

\subsection{An Illustrating Example}
\begin{figure}[tbp]
\centering\includegraphics[scale=0.45]{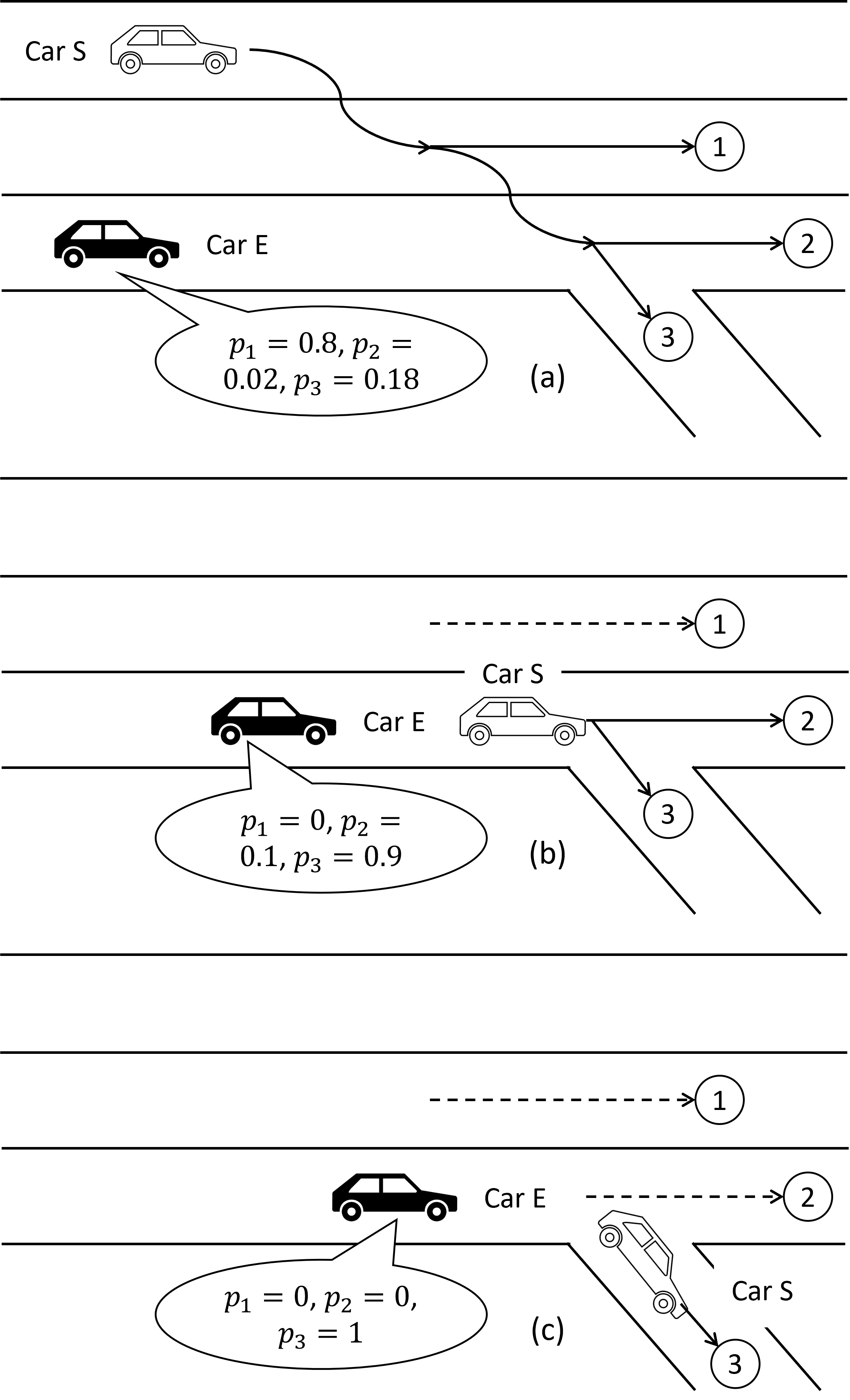}
\caption{The subplots (a), (b) and (c) show that the system states and the prediction for the surrounding vehicle change as time goes on. Solid lines and dash lines represent possible and impossible routes at that time. Note that the surrounding vehicle's states in (b) do not match the route prediction with the highest probability in (a), but our planner can still ensure system safety as it considers all possible predicted behaviors and trajectories.}
\label{fig:intuition}
\end{figure}

As stated in Section~\ref{sec: intro}, we consider lane changing in multi-lane highway as a representative application. 
Fig.~\ref{fig:intuition} shows an example scenario where the prediction for the surrounding vehicle gets adapted to the system states and the ego vehicle adjusts its action accordingly. Initially, as shown in the subplot (a), when the two vehicles are still far away from the highway off-ramp, the ego vehicle predicts that the probabilities of different routes for the surrounding vehicle are $p_1=0.8$, $p_2=0.02$ and $p_3=0.18$. A few seconds later, as shown in subplot (b), the surrounding vehicle is changing to the right-most lane before the off-ramp. The prediction gets adapted as the route $1$ is no longer possible. Finally, as shown in subplot (c), the surrounding vehicle exits from the off-ramp, and route $3$ becomes the only possible one.

We assume that the reward function measuring system performance is the average speed of the ego vehicle. Our planner behaves as follows. During (a), the ego vehicle may keep its speed or accelerate since route $1$ has the highest probability for the surrounding vehicle, as long as it is impossible to collide even in the worst case under all three possible routes. 
During (b), the ego vehicle remains safe because the less possible scenario (b) was considered in the planning process during (a). The ego vehicle will not necessarily decelerate harshly because route $3$ is the most possible one. During (c), the ego vehicle finds that the surrounding vehicle indeed took the most possible route $3$ in the prediction during (b), and can now accelerate without hesitance. 

Our speculative planning method maximizes the expected reward, while leaving enough buffer space to ensure safety for those less possible predicted scenarios, as shown below.


\subsection{Problem Formulation}
We denote the system state with $\mathop{\mathbb{S}}=\{ d_E, d_S, v_E, v_S, l_E, l_S \}$, where $d$, $v$ and $l$ denote the traveled distance, the velocity of the vehicle, and the lane that the vehicle is in, respectively. The subscript $E$ and $S$ of these variables correspond to ego vehicle and surrounding vehicle, respectively.

The probabilistic prediction of the surrounding vehicle's future trajectory can be represented as 
\begin{equation} \label{eq:prediction}
\mathop{\mathbb{P}}=\{\{ r_i, p_i, f_i(w_i)\}, i=1,2,\cdots,N\},
\end{equation}
where $r_i$ denotes the discrete route choice, $p_i$ denotes the probability of the route choice, $w_i$ denotes the vector of related parameters to define a trajectory corresponding to the route $r_i$, and $f_i(w_i)$ defines the probability distribution of $w_i$ under the route $r_i$. Assume that there are $N$ different possible route choices, we have 
\begin{equation} \label{eq:route_probability}
\sum _{i=1}^N p_i = 1.
\end{equation}

We assume that the prediction is conservative such that the real trajectory of the surrounding vehicle is always included and bounded by the prediction. The real trajectory is represented with $\{ \hat{r}, \hat{w}\}$, which are sampled from random variables $r$ and $w$ given $\mathop{\mathbb{P}}$. We have 
\begin{equation} \label{eq:real_traj}
\hat{r}=r_k
\text{, }
f_k(\hat{w})>0
\text{, }
\exists k \in \{1,2,\cdots,N\}.
\end{equation}

The system dynamics can be formulated as:
\begin{equation} \label{eq:dynamics}
\begin{cases}
\dot d_E(t)=v_E(t),
\\
\dot v_E(t)=u(t),
\\
l_E(t) \equiv l_E(t_0),
\\
\dot d_S(t)=v_S(t),
\\
\dot v_S(t)=\phi(\hat{w}),
\\
l_S(t)=\psi(\hat{r},\hat{w},d_S(t)),
\end{cases}
\end{equation}
where $u(t)$ is the control input, representing the acceleration for the ego vehicle $E$. $t_0$ is the initial time, and $l_E(t) \equiv l_E(t_0)$ means that the ego vehicle will go straight and stay in the lane. $\phi(\hat{w})$ is a function to derive acceleration of surrounding vehicle $S$ from the parameter vector $\hat{w}$. The function $\psi(\hat{r},\hat{w},d_S(t))$ can determine the lane that the surrounding vehicle is in. As shown in Fig.~\ref{fig:scenario}, we assign ids $\{0, 1, 2, 3\}$ to lanes from the leftmost to the rightmost, and lane $3$ corresponds to the off-ramp.

To ensure safety, the system need to satisfy 
\begin{equation} \label{eq:safety}
\begin{split}
l_E(t) \neq l_S(t) \text{ } \vee \text{ } |d_E(t)-d_S(t)| \geq d_{m} \text{, }\forall t \geq t_0
\\
\forall \hat{r}, \hat{w}\text{, s.t. Eq.~\eqref{eq:real_traj}},
\end{split}
\end{equation}
where $d_m$ is the minimum distance gap between vehicles to prevent collisions. Let $J(t)$ denote the reward function\footnote{$J(t)$ is the short form of $J(d_E(t), d_S(t), v_E(t), v_S(t), l_E(t), l_S(t), u(t))$.}. The sum of reward function over horizon $t \in [t_0, t_h]$, $\sum_{t=t_0}^{t_h} J(t)$, can be transferred to another function $Q(u(t_0), \bar{u}, \hat{r}, \hat{w})$\footnote{$\bar{u}$ represents a series of control inputs excluding $u(t_0)$, i.e., $u(t_0+\delta t : \delta t : t_h)$.}, based on Eq.~\eqref{eq:dynamics}. Since $\hat{r}$ and $\hat{w}$ are unknown,
our goal is to maximize the expectation of $Q(u(t_0), \bar{u}, r, w)$ given the probabilistic prediction $\mathop{\mathbb{P}}$. Then we have
\begin{equation} \label{eq:control_input}
\begin{split}
u(t_0)=\argmax_{u(t_0)} \mathop{\mathbb{E}_{ (r, w) \in \mathbb{P}}} \left[\max_{\bar{u}}(Q(u(t_0), \bar{u}, r, w))\right]
\\
\text{s.t. Eq.~\eqref{eq:safety}}.
\end{split}
\end{equation}
Similar to the receding horizon method in the Model Predictive Control, we will use only $u(t_0)$ for the current time step, and run the optimization in Eq.~\eqref{eq:control_input} periodically for the following control inputs.


\subsection{Speculative Planning with Adaptive Prediction}
Next we present our speculative planning algorithm, as shown in Algorithm~\ref{Speculative planning}, which generates control input for maximizing the expected reward. First, we set the initial value of $u(t)$ to be $0$, representing no acceleration (line 1). Similarly, we set the initial values for reward $\Omega$, safety indicator $\Theta$ and the minimum distance gap between vehicles $\delta d$ (lines 2-4).
Both $\Theta$ and $\delta d$ are acquired by the safety evaluation algorithm, as shown in Algorithm~\ref{Safety evaluation}. $\Theta$ is a binary variable, $\Theta = 0$ denotes that the system is safe now and there exists a series of control inputs $\bar{u}$ to keep the system safe, $\Theta = 1$ denotes that the system is possible to be unsafe given $\mathop{\mathbb{P}}$. Then let the temporary variable $a_t$ loop through the acceleration range $[a_{min}, a_{max}]$ (lines 5-6). We evaluate the system safety (line 7) and compute the expected reward given the probabilistic prediction $\mathop{\mathbb{P}}$ (line 8) for each $a_t$. If we get safety assurance with current action $a_t$ (line 9), or we get a higher reward and a larger distance gap (line 14), we update the control input $u(t)$ and its corresponding reward, safety indicator and minimum distance gap. $\varphi(\Omega_t, \delta d_t)$ is a function to balance reward and minimum gap, and a larger value is preferred. We will prove the safety guarantee of our algorithm (under certain conditions) later in Section~\ref{sec: safety proof}.

\begin{algorithm}[t]
\SetAlgoLined
\KwResult{Control input $u(t)$}
\SetKwInOut{Input}{Input}
\Input{$\mathop{\mathbb{S}}(t)=\{d_E(t), d_S(t), v_E(t), v_S(t), l_E(t), l_S(t)\}$, $\mathop{\mathbb{P}}=\{\{ r_i, p_i, f_i(w_i)\}, i=1,2,\cdots,N\}$}
$u(t) \leftarrow 0.0$\;
$\text{Reward } \Omega \leftarrow 0$\;
$\text{Safety indicator } \Theta \leftarrow 1$\;
$\text{Minimum gap } \delta d \leftarrow 100$\;
$a_t \leftarrow a_{min}$\;
\While{$a_t \leq a_{max}$}{
$\Theta_t, \delta d_t \leftarrow SafetyEval(\mathop{\mathbb{S}}(t), \mathop{\mathbb{P}}, a_t)$\;
$\Omega_t \leftarrow ExpectedReward(\mathop{\mathbb{S}}(t), \mathop{\mathbb{P}}, a_t)$\;
\uIf{$\Theta == 1 \text{ } \&\& \text{ } \Theta_t == 0$}{
$u(t) \leftarrow a_t$\;
$\Omega \leftarrow \Omega_t$\;
$\Theta \leftarrow \Theta_t$\;
$\delta d \leftarrow \delta d_t$\;
}
\ElseIf{$\Theta == 0 \text{ } \&\& \text{ } \Theta_t == 0 \text{ } \&\& \text{ } \varphi(\Omega_t , \delta d_t) \geq \varphi(\Omega , \delta d) $}{
$u(t) \leftarrow a_t$\;
$\Omega \leftarrow \Omega_t$\;
$\delta d \leftarrow \delta d_t$\;
}
$a_t \leftarrow a_t + \delta a$\;}
\caption{Speculative planning}
\label{Speculative planning}
\end{algorithm}

Safety evaluation is conducted as shown in Algorithm~\ref{Safety evaluation}. We initially assume that it is safe (line 1) and set the minimum distance gap to be $100$ (line 2). Then for every possible route (line 3), we assess whether it is still feasible according to the latest status of the surrounding vehicle $S$ (line 4). For example, if the surrounding vehicle is already on the off-ramp to exit the highway, routes 1 and 2 become impossible. We compute the union set of all possible future spatial-temporal trajectories~\cite{hong2017recognizing} of the surrounding vehicle, $\mathop{\mathbb{T}}_i$, even if the probability is small according to $f_i(w_i)$ (line 5). Assume that the ego vehicle adopts the control input $a_t$ at the current time step, and it is allowed to take any acceleration in the range $[a_{min}, a_{max}]$ for the following steps to prevent overlap with $\mathop{\mathbb{T}}_i$. We compute the minimum distance gap between $\mathop{\mathbb{T}}_i$ and the future trajectory of ego vehicle, $\delta d_{min}$ (line 6). We will update $\delta d_t$ with $\delta d_{min}$ acquired under route $r_i$ (line 7). If the gap is even less than the threshold $\delta d_s$, it is unsafe (lines 10-12). As stated before, this result is used in Algorithm~\ref{Safety evaluation} (line 7).

\begin{algorithm}[t]
\SetAlgoLined
\KwResult{Safety indicator $\Theta_t$, minimum gap $\delta d_t$}
\SetKwInOut{Input}{Input}
\Input{$\mathop{\mathbb{S}}(t)=\{d_E(t), d_S(t), v_E(t), v_S(t), l_E(t), l_S(t)\}$, $\mathop{\mathbb{P}}=\{\{ r_i, p_i, f_i(w_i)\}, i=1,2,\cdots,N\}$, $a_t$}
$\Theta_t \leftarrow 0$\;
$\delta d_t \leftarrow 100$\;
\For{$i \in \{1,2,\cdots,N\}$}{
\If{$\text{IsFeasible}(r_i, d_S(t), l_S(t))$}{
$\mathop{\mathbb{T}}_i \leftarrow Traj(d_S(t), v_S(t), l_E(t), l_S(t), r_i, f_i(w_i))$\;
$\delta d_{min} \leftarrow MinGap(\mathop{\mathbb{T}}_i, d_E(t), v_E(t), a_t)$\;
$\delta d_t \leftarrow min(\delta d_t, \delta d_{min}) $\;
}
}
\If{$\delta d_t \leq \delta d_s$}{
$\Theta_t \leftarrow 1$\;
}
\caption{\emph{SafetyEval(): Safety evaluation}}
\label{Safety evaluation}
\end{algorithm}

Algorithm~\ref{expected reward} presents the computation of the expected reward given $\mathop{\mathbb{P}}$. Let $\Omega_t$ and $p_t$ denote the sum of weighted reward and the sum of weights (i.e., probabilities), respectively. We assign initial values for them (lines 1-2). Similarly to Algorithm~\ref{Safety evaluation}, we loop through each feasible route $r_i$ (lines 3-4) and adapt our prediction and probability distribution according to the latest status of the surrounding vehicle (line 5). For example, assume that there is an appropriate road segment for the lane changing, as the surrounding vehicle moves forward towards the end of the road segment, the range of possible lane-changing positions can be smaller, thus the weight (i.e., probability) for the certain route needs to be adjusted. Under the specific route $r_i$, we sample the vector of parameters $w_{i,k}$ to generate the trajectory of the surrounding vehicle for $N_s$ times (lines 7-8). We compute the reward of the ego vehicle for each sample $\{r_i, w_{i,k}\}$, and $\Omega_{t,i}$ is the sum of all rewards (line 9). As $\Omega_{t,i} / N_s$ is the averaged reward under route $r_i$, we update $\Omega_t$ and $p_t$ accordingly (lines 11-12). Finally, $\Omega_t$ is scaled to be the expected reward (line 15), which as stated before, is used in Algorithm~\ref{Speculative planning} (line 8).

\begin{algorithm}[t]
\SetAlgoLined
\KwResult{Reward $\Omega_t$}
\SetKwInOut{Input}{Input}
\Input{$\mathop{\mathbb{S}}(t)=\{d_E(t), d_S(t), v_E(t), v_S(t), l_E(t), l_S(t)\}$, $\mathop{\mathbb{P}}=\{\{ r_i, p_i, f_i(w_i)\}, i=1,2,\cdots,N\}$, $a_t$}
$\Omega_t \leftarrow 0$\;
$p_t \leftarrow 0$\;
\For{$i \in \{1,2,\cdots,N\}$}{
\If{$\text{IsFeasible}(r_i, d_S(t), l_S(t))$}{
$p_i^{'}, f_i^{'}(w_i) \leftarrow Adapt(d_S(t), l_S(t), r_i, p_i, f_i(w_i))$\;
$\Omega_{t,i} \leftarrow 0$\;
\For{$k \in \{1,2,\cdots,N_s\}$}{
$w_{i,k} \leftarrow Sample(f_i^{'}(w_i))$\;
$\Omega_{t,i} \leftarrow \Omega_{t,i} + \max_{\bar{u}}(Q(a_t, \bar{u}, r_i, w_{i,k}))$\;
}
$\Omega_t \leftarrow \Omega_t + p_i^{'} \Omega_{t,i}/N_s$\;
$p_t \leftarrow p_t + p_i^{'}$\;
}
}
$\Omega_t \leftarrow \Omega_t / p_t$\;
\caption{\emph{ExpectedReward(): Computation of the expected reward }}
\label{expected reward}
\end{algorithm}


The prediction adaptation is based on two assumptions: (1) the surrounding vehicle does not move backward; (2) after a complete and safe lane changing, the surrounding vehicle does not return back to the original lane. When the prediction is outdated, we filter out those impossible behaviors and trajectories of the surrounding vehicle based on its latest status, and scale the probabilities of the remaining such that the sum is still $1$.

\subsection{Safety Guarantee}\label{sec: safety proof}

\begin{theorem}\label{proposition1}
For a dynamical system defined by Eq.~\eqref{eq:dynamics}, our proposed planner (Algorithms~\ref{Speculative planning}, \ref{Safety evaluation} and~\ref{expected reward}) will ensure system safety, if Eq.~\eqref{eq:real_traj} holds and there exists a safe planning decision at the initial state of the system.\end{theorem}

\begin{proof}
Let us prove this by contradiction. We first assume that there exists a time $t_s$ such that $\Theta(t_s)=0$ and $\Theta(t_s+\delta t)=1$. 
According to Algorithm~\ref{Speculative planning}, if there exists $u(t_s+\delta t)$ such that $\Theta_t(t_s+\delta t)=0$, then $\Theta(t_s+\delta t)=0$. So we have $\Theta_t(t_s+\delta t)=1$, $\forall u(t_s+\delta t)$. According to Algorithm~\ref{Safety evaluation}, there exists at least one feasible route $r_i$ such that $\delta d_{min}(t_s+\delta t) \leq \delta d_s < \delta d_{min}(t_s)$, $\forall u(t_s+\delta t)$. Since $d_E(t_s+\delta t)$ and $v_E(t_s+\delta t)$ are acquired by substituting $u(t_s)$ into Eq.~\eqref{eq:dynamics}, there exists an action $a_t(t_s+\delta t)$ such that $\delta d_{min}(t_s+\delta t) = MinGap(\mathop{\mathbb{T}}_i(t_s+\delta t), d_E(t_s+\delta t), v_E(t_s+\delta t), a_t(t_s+\delta t)) = MinGap(\mathop{\mathbb{T}}_i(t_s+\delta t), d_E(t_s), v_E(t_s), a_t=u(t_s))$. Thus, $MinGap(\mathop{\mathbb{T}}_i(t_s+\delta t), d_E(t_s), v_E(t_s), a_t=u(t_s))<\delta d_{min}(t_s)=MinGap(\mathop{\mathbb{T}}_i(t_s), d_E(t_s), v_E(t_s), a_t=u(t_s))$. It requires $\mathop{\mathbb{T}}_i(t_s+\delta t) \setminus \mathop{\mathbb{T}}_i(t_s) \neq \emptyset$. However, $\mathop{\mathbb{T}}_i$ is the union set of all possible future spatial-temporal trajectories of the surrounding vehicle, $\mathop{\mathbb{T}}_i(t_s+\delta t) \subseteq \mathop{\mathbb{T}}_i(t_s)$. From this contradiction, we know that $\Theta \equiv 0$ and the system is safe under our proposed planner.
\end{proof}


\subsection{Surrounding Vehicle's Model}
It is worth noting that the generality of our planning method will not be affected by the model of the surrounding vehicle. 

For longitudinal motion, we assume that the surrounding vehicle is controlled to maintain the desired speed $v_d$ under the acceleration function $\phi(\hat{w})$. For lateral motion, we assume that the surrounding vehicle indicates a right turn and intends to change lanes when $d_S=d_{lc}^0$. The route is randomly determined according to $p_1$, $p_2$ and $p_3$. Route~$1$ corresponds to change lane once, while route $2$ and $3$ need to change lanes twice. Based on the selected route, the execution of first and second lane changing happen at $d_{lc}^1$ and $d_{lc}^2$, respectively. We assume that $d_{lc}^1-d_{lc}^0$ and $d_{lc}^2-d_{lc}^1$ are closely related to the personality of the driver, which is represented by aggressiveness~\cite{yu2018human,liu2020impact}. We then have
\begin{equation} \label{eq:aggressive}
\delta d_{lc}= d_a q_a + d_c + d_n,
\end{equation}
where $q_a$, the aggressiveness level, is a real variable satisfying $-1 \leq q_a \leq 1$. A larger $q_a$ represents a more aggressive driver. $d_a < 0$ is the coefficient for the aggressiveness term, and $d_c$ is a constant. With the noise term $d_n$, $d_{lc}^1-d_{lc}^0$ and $d_{lc}^2-d_{lc}^1$ are not necessarily the same. For a more aggressive surrounding vehicle, $d_{lc}^1-d_{lc}^0$ and $d_{lc}^2-d_{lc}^1$ are smaller, which leaves less time for the ego vehicle to react.

\section{Experimental Results}\label{sec: experiment}

\subsection{Effectiveness of Our Approach}

We compare system safety and performance under different planners to demonstrate the strength of our approach: baseline methods `$\text{IDM}_1$', `$\text{IDM}_2$' and `$\text{IDM}_3$' are all based on Intelligent Driver Model~\cite{alhariqi2022calibration,jin2016optimal}, which is a common car following model in the transportation domain. `$\text{IDM}_1$' is the original version that the ego vehicle only follows the surrounding vehicle in the same lane, `$\text{IDM}_2$' means that the ego vehicle will follow the surrounding vehicle that intends to change lanes from the adjacent lane as well, `$\text{IDM}_3$' means that the ego vehicle follows the surrounding vehicle in any of the three lanes in the highway. A larger subscript corresponds to earlier reactions to the surrounding vehicle, which is expected to be safer and less efficient. Baseline method `MPC' represents the Model Predictive Control approach~\cite{liu2021trajectory,xiao2022robotic,9712223}, which can address the uncertainty in the behavior and trajectory of the surrounding vehicle. However, it does not consider the underlying probability distribution. `SPAP' is our proposed method, Speculative Planning with Adaptive Prediction. For the reward function, without losing generality, we assume that $J(t)=v_E(t)$. 'MPC+agg' and 'SPAP+agg' are extensions of MPC and our method with consideration of vehicle aggressiveness.

We set the horizon of each simulation to be $12$ seconds, and the simulation step size and control step size $\delta t$ are both at $0.1$ second. The desired speed is $v_d=25$ m/s, and the speed limit is $30$ m/s in the highway. We randomly sample the values for $p_1$, $p_2$ and $p_3$ and evaluate the performance of different planners with or without the predicted aggressiveness of the surrounding vehicle, as shown in Table~\ref{table: statistical evaluation}. Each row corresponds to the averaged results of $10,000$ simulations. We can see that \textbf{our SPAP and SPAP+agg planners provide significant improvement on system performance (larger average speed) over the baseline planners, while ensuring system safety}. SPAP+agg and MPC+agg provide improvement over SPAP and MPC, as we assume that the aggressive prediction is accurate and can help reduce the uncertainty in predicting surrounding vehicle behavior (if not, they may not provide such improvement).


\begin{table}[h]
\centering
\caption{Safety and performance evaluation for different planners with or without predicted aggressiveness of the surrounding vehicle.
}
\label{table: statistical evaluation}
\scalebox{1.0}{
\begin{tabular}{|c|c|c|c|}
\hline
Planners            & safety rate & average speed & final speed \\ \hline
$\text{IDM}_1$      &     94.73\%      &     24.73 m/s &    24.68 m/s\\ \hline
$\text{IDM}_2$      &     96.57\%      &     23.76 m/s &    23.38 m/s\\ \hline
$\text{IDM}_3$      &     100\%        &     22.96 m/s &    23.82 m/s\\ \hline
MPC                 &     100\%       &     25.85 m/s &    29.21 m/s\\ \hline
MPC+agg             &     100\%        &     26.45 m/s &    29.26 m/s\\ \hline
\textbf{SPAP}                &     \textbf{100\%}        &    \textbf{27.56 m/s} &    \textbf{29.45 m/s}\\ \hline
\textbf{SPAP+agg}            &     \textbf{100\%}        &     \textbf{27.90 m/s} &    \textbf{29.53 m/s}\\ \hline
\end{tabular}
}
\end{table}

\begin{figure}
    \centering
    \includegraphics[width=0.52\textwidth]{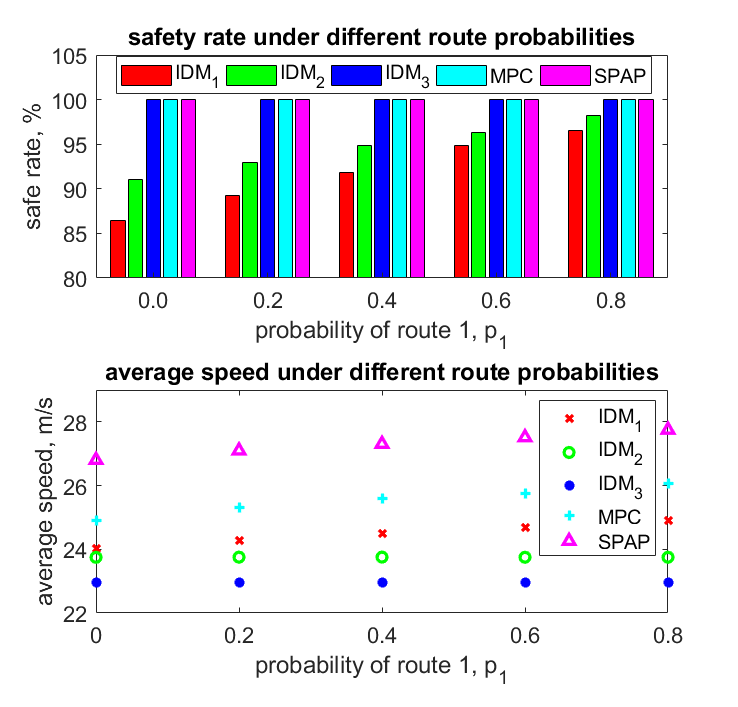}
    \caption{Safety rate and average speed of the ego vehicle under different planners are compared when the probability of route $1$, $p_1$, changes. The probabilities of the other two routes are set as $p_2=0.8-p_1$ and $p_3=0.2$.}
    \label{fig: experiment1}
\end{figure}

We conduct additional experiments to further study the impact of different route probabilities. 
Fig.~\ref{fig: experiment1} presents the safety rate and average speed of the ego vehicle when the probability of route $1$, $p_1$, changes. We set $p_3=0.2$, and $p_2$ is determined such that $p_1 + p_2 + p_3 =1$. 
As we expected, since it is route $2$ that has the most interference with the ego vehicle, the safety rate is the lowest when $p_1=0$ and $p_2=0.8$ for planners $\text{IDM}_1$ and $\text{IDM}_2$. The safety rate increases when $p_1$ increases gradually. For planners $\text{IDM}_3$, MPC and SPAP, there is no any collision. $\text{IDM}_3$ is over-conservative by its nature, MPC is formulated with safety constraints, and our proposed SPAP has safety guarantee, as presented in Section~\ref{sec: safety proof}.

Fig.~\ref{fig: experiment1} also shows that the average speed increases slightly as $p_1$ increases. For the three IDM-based planners, it meets our expectation that a larger subscript corresponds to a less efficient planner, thus resulting in a lower average speed. MPC performs better than these IDM-based planners. 

\begin{table}[]
\centering
\caption{Computation time, safety and performance evaluation under different sampling times $N_s$.
}
\label{table: sampling and computation time}
\scalebox{1.0}{
\begin{tabular}{|c|c|c|c|c|}
\hline
$N_s$   & time cost & safety rate & average speed & final speed \\ \hline
10      &    0.03 s &     100\%   &     26.21 m/s &    28.39 m/s\\ \hline
25      &    0.04 s &     100\%   &     26.82 m/s &    28.47 m/s\\ \hline
50      &    0.07 s &     100\%   &     27.30 m/s &    28.64 m/s\\ \hline
100     &    0.12 s &     100\%   &     27.31 m/s &    28.63 m/s\\ \hline
200     &    0.22 s &     100\%   &     27.27 m/s &    28.64 m/s\\ \hline
\end{tabular}
}
\end{table}


\subsection{Real-Time Computation Complexity}
Since the control step size is set to $0.1$ second, the planner could only be employed in real-time if the computation time is within $0.1$ second. To evaluate the real-time computation complexity of our approach, we set $p_1=0.4$, $p_2=0.4$, $p_3=0.2$ and conduct experiments in a server with Intel(R) Xeon(R) Gold 6130 CPU @ 2.10GHz. We choose the number of sampling times $N_s$ to be $10$, $25$, $50$, $100$ and $200$. Intuitively, the larger the $N_s$, the more accurately the planner samples the prediction distribution and the better the performance, however at a higher computational cost. 

From the results in Table~\ref{table: sampling and computation time}, we can see that when $N_s$ increases from $10$, $25$ to $50$, there is a substantial increase of system performance (i.e., average speed the vehicle can safely achieve). Further increasing $N_s$ to $100$ and $200$ will not lead to significant change in performance. 
Thus, $N_s=50$ seems to be the sweet spot that balances performance and computational load for this application. Note that the real-time computation demand can be satisfied when $N_s  = 50$, as the time cost of $0.07$ second is smaller than $0.1$ (moreover, the reported average computation time includes not only the time spent on planning, but also the time on the simulator; the real computation time is lower). 
We plan to conduct more evaluations on other driving tasks and computing platforms in the future work.


\section{Conclusion}\label{sec: conclusion}
We presented a speculative planning framework to ensure safety and improve performance in uncertain and ambiguous traffic scenarios.
By adapting the prediction results according to the system states, impossible behaviors and trajectories of the surrounding vehicle are filtered out, thus leading to more effective planning. Through the case study of lane changing in a multi-lane highway, we demonstrate the advantages of our proposed planner over various baseline methods. 





\bibliographystyle{IEEEtran}
\bibliography{Refs}

\end{document}